\documentclass[10pt,twocolumn,letterpaper]{article}

\usepackage{wacv}
\usepackage{times}
\usepackage{epsfig}
\usepackage{graphicx}
\usepackage{amsmath}
\usepackage{amssymb}
\usepackage{amsfonts} 
\usepackage{booktabs}
\usepackage{multirow}
\usepackage{bm}
\usepackage{array}
\usepackage[bookmarks=false]{hyperref}

\newcolumntype{C}[1]{>{\centering\let\newline\\\arraybackslash\hspace{0pt}}m{#1}}

\wacvfinalcopy
\ifwacvfinal\fi

\begin{document}

\title{ULSAM: Ultra-Lightweight Subspace Attention Module for Compact Convolutional Neural Networks}

\author{Rajat Saini \textsuperscript{*} \\
IIT Hyderabad\\
{\tt\small cs17mtech11002@iith.ac.in}
\and
Nandan Kumar Jha   \thanks{The  first two authors contributed equally to this work.} \hspace{0.01cm} \thanks{Corresponding author.}\\
IIT Hyderabad\\
{\tt\small cs17mtech11010@iith.ac.in}
\and
Bedanta Das \\
IIT Hyderabad\\
{\tt\small cs17mtech11009@iith.ac.in}
\and
Sparsh Mittal \\
IIT Roorkee\\
{\tt\small sparshfec@iitr.ac.in}
\and
C. Krishna Mohan \\
IIT Hyderabad\\
{\tt\small ckm@iith.ac.in}
}

\maketitle
\ifwacvfinal\thispagestyle{empty}\fi
\begin{abstract}
The capability of the self-attention mechanism to model the long-range dependencies has catapulted its deployment in vision models. Unlike convolution operators, self-attention offers infinite receptive field and enables compute-efficient modeling of global dependencies. However, the existing state-of-the-art attention mechanisms incur high compute and/or parameter overheads, and hence unfit for compact convolutional neural networks (CNNs). In this work, we propose a simple yet effective ``Ultra-Lightweight Subspace Attention Mechanism'' (ULSAM), which infers different attention maps for each feature map subspace. We argue that leaning separate attention maps for each feature subspace enables multi-scale and multi-frequency feature representation, which is more desirable for fine-grained image classification. Our method of subspace attention is orthogonal and complementary to the existing state-of-the-arts attention mechanisms used in vision models. ULSAM is end-to-end trainable and can be deployed as a plug-and-play module in the pre-existing compact CNNs. Notably, our work is the first attempt that uses a subspace attention mechanism to increase the efficiency of compact CNNs.  To show the efficacy of ULSAM, we perform experiments with MobileNet-V1 and MobileNet-V2 as backbone architectures on  ImageNet-1K  and three fine-grained image classification datasets. We achieve  $\approx$13\% and $\approx$25\% reduction in both the FLOPs and parameter counts of MobileNet-V2 with a 0.27\% and more than 1\% improvement in top-1 accuracy on the ImageNet-1K and fine-grained image classification datasets (respectively). Code and trained models are available at \url{https://github.com/Nandan91/ULSAM}.

\end{abstract}

\section{Introduction}

Convolutional neural networks (CNNs) have achieved remarkable predictive performance in various cognitive and learning tasks \cite{2019_TPAMI_GaoRes2Net,Xie_2017_CVPR,He_2016_CVPR,Simonyan2014VeryDC,NIPS_2012_Krizhevsky}. The unprecedented predictive performance of CNNs stems from the rich representational power of CNNs, which in turn stems from the deeper and wider layers in networks. Deeper and wider layers enhance the expressiveness and discrimination ability of the network by circumventing the inherent limitations of convolution operators, viz., locality \cite{NIPS_2018_A2NET} and linearity \cite{2013_Lin_NiN}.

The locality of the seminal operator, convolution, in CNNs offers a theoretical guarantee, unlike the shallow networks, to avoid the {\em curse of dimensionality} while approximating the hierarchically local compositional functions \cite{Poggio2016WhyAW,2016_arXiv_Mhaskar}.  Since convolution in CNN capture the local (e.g., $3\times3$) feature correlations \cite{1998_IEEE_Lecun}, multiple convolution operators are stacked together to enlarge the effective receptive field size and capture the long-range dependencies \cite{NIPS_2018_A2NET}. However, this makes the CNNs {\em deeper}.  Further, since the linearity of convolution operation leads to inefficient capturing of the non-linear abstractions in input data \cite{2013_Lin_NiN}, CNNs employ a higher number of filters per layer, which are learned to capture all the possible variations of complex latent concept  \cite{2013_Lin_NiN}.  However, this makes the CNNs {\em wider}.  Deeper and wider layers in CNNs \cite{He_2016_CVPR,Xie_2017_CVPR,2016_ECCV_ResNetV2} leads to a high computational cost (measured in the number of floating-point operations or FLOPs) and a large number of parameters which makes deployment of CNNs on resources-constrained platforms quite challenging.

The compact CNNs such as MobileNets \cite{Howard2017MobileNetsEC,2018_CVPR_Sandler} and ShuffleNets \cite{Zhang_2018_CVPR,Ma_2018_ECCV} seek to reduce the computational cost significantly by employing depthwise separable (DWS) convolution. Similarly, the dilated convolution has been employed to enlarge the receptive field size in vision tasks \cite{2016_iCLR_Yu}. However, the inefficiencies of convolution operators still exist, and the network learns complex cross channel inter-dependencies in a computationally-inefficient manner.

The success of self-attention in natural language processing \cite{NIPS_2017_Vaswani} in modeling the long-range dependencies has enabled its inclusion as a computation primitive in vision models \cite{2019_NIPS_Ramachandran,2019_ICCV_Bello,NIPS_2018_A2NET}. Self-attention efficiently captures the global dependencies of features in feature space and circumvents the inherent limitations of convolution operators in CNNs. However, the higher parameters and/or computation overheads of these attention mechanisms (Table \ref{tab:ComparisonOfAttentions}) are undesirable for in compact CNNs. 
Since redundancy in the parameter space of compact CNNs is low, the desirable attention mechanism for compact CNNs should have the capability to capture the global correlation (fusing semantic and spatial information) more effectively and efficiently compared to the existing attention mechanism.

\begin{figure}[htbp]
\centering

\includegraphics[scale=0.45]{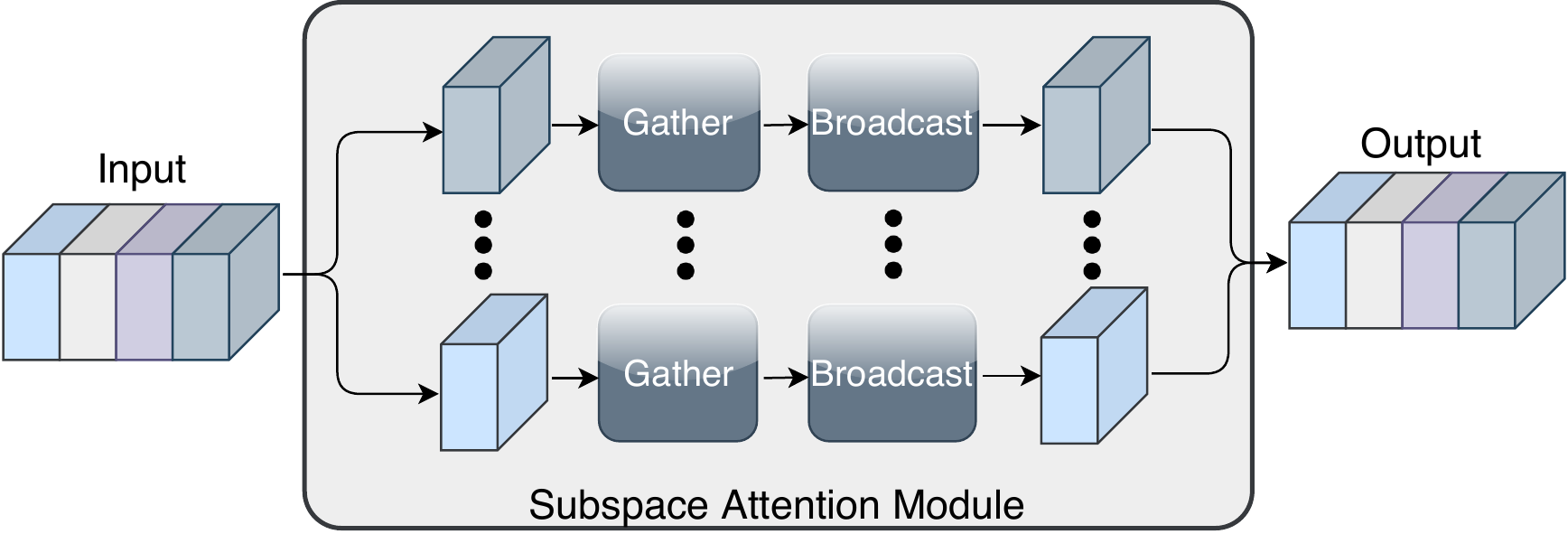}
\caption{A block diagram of ULSAM}
\label{fig:IntroULSAM}
\end{figure}

In this work, we propose the ``Ultra-Lightweight subspace attention module'' (ULSAM), a novel attention block for compact CNNs (Figure \ref{fig:IntroULSAM}). ULSAM learns different attention maps for each feature map subspace and reduces the spatial and channel redundancy in feature maps of compact CNNs. Also, learning different attention maps for each subspace enables multi-scale and multi-frequency feature representation, which is desirable, especially for the fine-grained image classification tasks. To the best of our knowledge, ULSAM is the first attention module which enables efficient (compute and parameter) learning of cross channel inter-dependencies in each subspace of feature maps.

Our key contributions are summarized as follows: 

\begin{itemize}
\item We propose a novel attention block (ULSAM) for compact CNNs, which learns individual attention maps for each feature subspace and enables compute-efficient learning of cross-channel information along with multi-scale and multi-frequency feature learning. 
\item We show the effectiveness of our block (ULSAM) through extensive experiments with MobileNet-V1 and MobileNet-V2 on ImageNet-1K as well as three fine-grained image classification datasets. 
\item  We demonstrate the scalability and efficacy of ULSAM with fewer parameters and computations on MobileNet-V1 and MobileNet-V2 with ImageNet-1K and fine-grained datasets.
\end{itemize}

\section{Background and Related Work}
This section describes various design techniques, along with their limitations, which have been incorporated in CNNs to reduce the number of FLOPs. Further, we brief state-of-the-art attention mechanisms employed in vision models for reducing the FLOPs and parameter counts by virtue of feature re-distribution in the feature space. We show the overhead, in terms of FLOPs and parameters counts, of these attention mechanisms along with our proposed module, ULSAM, in Table \ref{tab:ComparisonOfAttentions}.

\subsection{Computational Cost of Convolution in CNNs}

\begin{table*}[htbp]
\centering
\caption{Compute and parameter overheads of different attention modules. These overheads are compared with our proposed attention module ``ULSAM" (assuming $m$ = 512, $t$ = $\frac{m}{8}$, $r$ = 16, $h\times w$ = $14\times 14$, and dilation rate is 4 in BAM \cite{2018_BMVC_Park}).}

\resizebox{0.99\textwidth}{!}{       
\begin{tabular}[htbp]{l| C{1.3cm}| c |C{1.1cm}| l | l | C{1.1cm}|C{1.1cm}|C{1.1cm}|C{1.1cm} }
\hline
Attention module & subspace attention& MLP & costlier $1\times1$ conv & \#Params &  \#FLOPs & \#Params ($\times10^3$) &  \#FLOPs ($\times10^6$)  &  \#Params $(norm.)$&  \#FLOPs $(norm.)$ \\ \hline
 
Non-local \cite{Wang_2018_CVPR} & $\times$ & $\times$ & \checkmark & $2m^2$& $2m^2 h w$& 524 & 102.76  & $512\times$ &$512\times$\\
 $ A^2$ - Net \cite{NIPS_2018_A2NET} & $\times$ & $\times$ & \checkmark & $2mt$& $2m t h w$& 66 & 12.85  & $64\times$& $64\times$\\
 SE-Net \cite{Hu_2018_CVPR} & $\times$ & \checkmark &$\times$ & $\frac{2m^2}{r}$ & $\frac{2m^2}{r}$& 33& 0.03 & $33\times$ & $0.16\times$\\
 BAM \cite{2018_BMVC_Park} & $\times$ & \checkmark & \checkmark & $\frac{4m^2}{r} + \frac{18m^2}{r^2}$ & $\frac{2m^2}{r} + (\frac{4m^2}{r} + \frac{18m^2}{r^2})hw$ & 84 & 16.49  & $82\times$& $82.16\times$ \\ 
 CBAM \cite{ECCV_2018_CBAM} & $\times$ & \checkmark & $\times$ & $\frac{2m^2}{r}$ + $98$ & $\frac{2m^2}{r}$ +  $98 h w$ & 33 & 0.05  & $33\times$& $0.26\times$ \\
 ULSAM ({\bf ours}) & \checkmark & $\times$ & $\times$ & $2m$& $2mh w$ & 1 & 0.2  & $1\times$& $1\times$\\
            
\hline
\end{tabular}}
\label{tab:ComparisonOfAttentions}      
\end{table*} 

  
The computational cost (in FLOPs) of standard convolution (SConv)  and FC layers of filter size $s_k\times s_k$ is shown in Eq. \ref{eqn:flops} , where  $m$ and $n$ are the number of input feature maps and output feature maps (respectively) of spatial size $h\times w$.  Note that the FC layer is a special case of the convolution layer where the filter size is equal to the input feature map size, and hence, the size of output feature maps size in the FC layer is $1\times 1$.

\begin{align} \label{eqn:flops}
 \text{FLOPs in SConv} = s_k\times s_k\times m\times n\times h\times w  
\end{align}

In state-of-the-art CNNs \cite{2018_CVPR_Sandler,Howard2017MobileNetsEC,He_2016_CVPR}, $m$ and $n$ are in the order of thousands (in deeper layers of CNNs) hence, the value of $m\times n$ is  substantially large. In Eq. \ref{eqn:flops}, the term $m\times n$ coupled with  $s_k\times s_k$ stems from the combined feature extraction and feature aggregation in SConv.  To reduce the computational cost depthwise separable (DWS) convolution \cite{2018_CVPR_Sandler,Howard2017MobileNetsEC,Chollet_2017_CVPR} has been deployed which decouples the $m\times n$ with $s_k\times s_k$ by decomposing the standard convolution into depthwise convolution (feature extraction) and pointwise convolution (feature aggregation). Depthwise convolution reduces redundancy in the channel extent and breaks the fully connected pattern between input and output feature. Similarly,  pointwise convolution reduces the redundancy in the spatial extent. 
The computational cost of DWS convolution is $s_k\times s_k\times m\times h\times w$ (depthwise conv) + $m\times n\times h\times w$ (pointwise conv). This decomposition in DWS convolution reduces the computations significantly; however, due to the $m\times n$ term in pointwise convolution, the computations in DWS convolution are dominated by the pointwise convolution. For example, in MobileNet-V1 \cite{Howard2017MobileNetsEC}, pointwise convolution accounts for 94.86\% of the total FLOPs while depthwise convolution accounts for only 3.06\% of the total FLOPs. 

To reduce the computation overhead of pointwise convolution, ShuffleNet \cite{Zhang_2018_CVPR} employs group convolution on $1\times 1$ layers. In group convolution, input feature maps are divided into mutually exclusive groups, and each group independently convolves with $s_k\times s_k$ filters and breaks the fully-connected patterns among input and output features maps. This reduces the number of computations; however, the stacking of multiple group convolution layers prohibits the cross-group interaction and leads to a drop in performance \cite{Zhang_2018_CVPR}.  To tackle this issue, ShuffleNet employs the channel shuffling mechanism after the group convolution layer. However, the channel shuffling does not scale with a larger number of groups, and with an increasing number of groups, channel shuffling gives diminishing returns \cite{Zhang_2018_CVPR}. Also, channel shuffling is naive and context-unaware, and hence, the limitations of group convolution are not adequately mitigated.  

ULSAM employs attention mechanism and enables compute-efficient interaction of cross-channel information where only one $1\times 1$ filter is used after depthwise convolution, i.e., $n$ =1 and there is no $m\times n$ term in the computation. Therefore, ULSAM decomposes the dense connections among input and output feature maps {\em without incurring the limitations of group convolution and computation overhead of pointwise convolution}.

\subsection{Attention Mechanisms for Vision Models}
In guided cognition tasks, attention is a way to assign different importance to different parts of the input data, which enables the networks to pick the salient information from the noisy data \cite{2015_ICML_Xu,2018_Wang_Attention}.  Attention can be broadly categorized into two categories viz. implicit attention and explicit attention. During the training, CNNs naturally learn a form of implicit attention where the neurons in CNN respond differently to different parts of input data \cite{olah2018the,olah2017feature,ECCV_2014_Zeiler}.

Recently, there has been a growing interest in incorporating explicit attention into CNNs for vision-related tasks.   Xu et al. \cite{2015_ICML_Xu} and Chen et al. \cite{Chen_2017_CVPR_SCA} use attention  mechanism to generate captions from the images. Wang et al. \cite{Wang_2017_CVPR_Residual_attention} propose a residual attention network by stacking multiple attention modules to generate attention-aware features. Several other recent works incorporate explicit attention mechanisms to improve the computational efficiency and feature distribution in feature space and improve the efficiency of CNNs.  Wang et al. \cite{Wang_2018_CVPR} proposed non-local operation, which is a generalized form of self-attention \cite{NIPS_2017_Vaswani},  to boost the performance in video recognition tasks. Chen et al. \cite{NIPS_2018_A2NET} introduced a double attention block that captures the long-range dependencies by gathering and distributing features in the entire feature space. Park et al. \cite{2018_BMVC_Park} proposed the ``Bottleneck Attention Module" (BAM), which employed a multi-layer perceptron (MLP) for channel-wise feature aggregation and dilated convolution for efficiently extracting the spatial information. Woo et al. \cite{ECCV_2018_CBAM} introduced the ``convolution block attention module'' (CBAM), which exploits both spatial and channel-wise feature correlation using attention mechanism to improve the representational power of CNNs. SE-Net \cite{Hu_2018_CVPR} employed MLP, which re-calibrates the feature maps through squeeze and excitation operations.  

To show the efficacy of the aforementioned attention mechanism for compact CNNs, we calculate the computation (FLOPs) and parameter (Params) overhead of deploying these attention mechanisms in CNNs (Table \ref{tab:ComparisonOfAttentions}). In Table \ref{tab:ComparisonOfAttentions},  $t$ is the number of attention maps in $A^2$ - Net and $r$ is  the reduction ratio (hyper-parameter) for MLP in BAM, CBAM, and  SE-Net.  As shown in  Table \ref{tab:ComparisonOfAttentions}, the existing attention mechanisms incur high computational overhead (due to costlier $1\times 1$ convolution used for generating attention maps) and/or parameter overhead (due to the parameter-heavy MLP). On the contrary, ULSAM uses only one $1\times 1$ filter and exploits the linear relationship between feature maps and avoids the use of MLP.

\section{Proposed Method}
The expressiveness of the self-attention layer with a sufficient number of attention heads and relative positional encoding is higher than that of the convolution layer \cite{2020_iCLR_Cordonnier}. However, the large number of attention heads can lead to high compute and parameter overheads, especially in the initial layers of CNNs, where the dimensions of feature maps are high.  For example, to replace a convolution layer with $7\times 7$ filters (in the initial layer) with a self-attention layer, at least 49 attention heads need to be deployed \cite{2020_iCLR_Cordonnier}. 

\begin{figure*}[htbp]\centering
\includegraphics[scale=0.65]{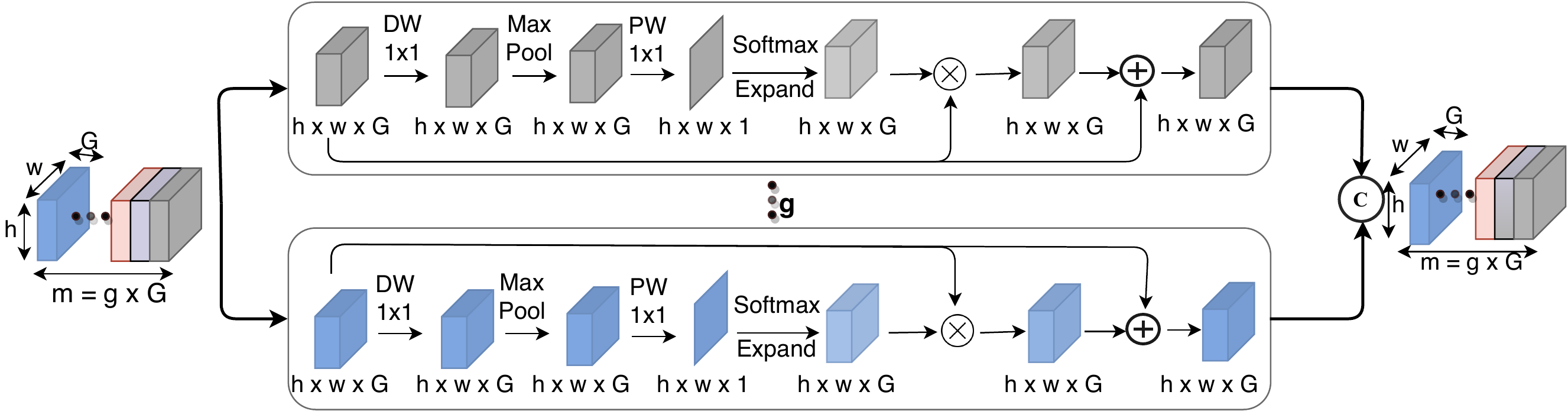}
\caption{ULSAM divides the input feature maps into $g$ mutually exclusive groups where each group contains $G$ feature maps.}
\label{fig:ULSAM}
\end{figure*}

\subsection{Design Optimization in ULSAM}

In ULSAM, we use only one attention map for each feature subspace. Further, unlike \cite{NIPS_2018_A2NET} and \cite{Wang_2018_CVPR}, we use depthwise convolution in the initial step and only one filter in pointwise convolution in the later step of generating the attention maps. This reduces the computations substantially and makes ULSAM suitable for compact CNNs.

{\bf Multi-scale feature representation:}
Multi-scale feature learning helps in understanding the context of the image, especially in a multi-object environment, and improve the representational power of networks \cite{2019_TPAMI_GaoRes2Net,2019_ICLR_ChenBigLiitle,2019_CVPR_WangElastic}. Moreover, multi-scale feature learning enables the network to encode the positional information in an efficient manner, which is desirable for position-dependent vision tasks such as semantic segmentation \cite{2020_ICLR_Islam}. Compared to employing the filters of different receptive size (for eg., InceptionNet variants \cite{Szegedy_2015_CVPR,Szegedy_2016_CVPR,Szegedy2017Inceptionv4IA}), dividing the feature maps into stages (feature map subspace) and applying different convolution for each stage is an efficient way of generating the multi-scale features which improve the effective receptive size of the networks \cite{2019_TPAMI_GaoRes2Net}. Moreover, the predictive performance of network increases with the increasing number of stages in feature map, and hence, compared to increasing depth, width, cardinality,  stages in feature maps is a more effective way to boost the representational power of networks \cite{2019_TPAMI_GaoRes2Net}. Therefore we divide the feature maps into different subspace and infer different attention maps for each subspace in ULSAM, which enables the multi-scale feature representation. 

{\bf Multi-frequency feature representation:}
The natural images composed of low frequency and high-frequency components where the former entails slowly varying features, and the latter represents the fine details in image \cite{2019_ICCV_chenOctave}. Imparting unequal importance to high and low-frequency features is an efficient way of representing features in feature maps \cite{2019_ICCV_chenOctave,2019_BMVC_Cheng}. Following this, learning different importance through different weights in different attention maps for each feature subspace is an efficient way of learning multi-frequency features. This way of learning multi-frequency features is beneficial when there are high intra-class variations present in image samples. Hence, multi-frequency features leaning is more desirable for fine-grained image classification where discriminative regions have fine details, i.e., high-frequency features. 

\subsection{ULSAM}

Let $F \in R^{m \times h \times w}$ be the feature maps from an intermediate convolution layer, where $m$ is the number of input channels,  $h$, and $w$ is the spatial dimensions of the feature maps. Our objective is to learn to capture the cross-channel inter-dependencies in the feature maps efficiently. As shown in Figure \ref{fig:ULSAM}, ULSAM divides the input feature maps ($F$) into $g$ mutually exclusive groups $[F_1, F_2,....F_{\tilde n},....F_g]$  where each group have $G$ feature maps.  We define  $F_{\tilde n}$ as a group of intermediate feature maps and proceed as follows.  
 \begin{align}
    A_{\tilde n} &=  softmax(PW^{1}(maxpool^{3 \times 3, 1}(DW^{1 \times 1}(F_{\tilde n})))) \label{eqn:AttentionMap}  \\
    \hat F_{\tilde n} &= (A_{\tilde n} \otimes F_{\tilde n})\oplus F_{\tilde n}  \label{eqn:Concatenation} \\
    \hat F &= concat([\hat F_1,\hat F_2,....\hat F_{\tilde n},....\hat F_{g}])  \label{eqn:ULSAMFinalOutput} 
\end{align}

In Eq. \ref{eqn:AttentionMap},  $maxpool^{3 \times 3, 1}$ is maxpool with kernel size = $3 \times 3$ and padding = 1, $DW^{1 \times 1}$ is depthwise convolution with  $1 \times 1$ kernel,  $PW^{1}$ is pointwise convolution with only one filter, and  $A_{\tilde n}$ is an attention map inferred from a group of intermediate feature maps ($F_{\tilde n}$). Attention map ($A_{\tilde n}$) in each group (subspace)  captures the non-linear dependencies among the feature maps by learning to gather cross channel information. To ensure  $\sum_{i,j} A_{\tilde n} (i,j) = 1$, i.e. a valid attention weighting tensor, we employ a gating mechanism with a softmax activation in Eq. \ref{eqn:AttentionMap}. Each group of feature maps gets the refined set of feature maps ($\hat F_{\tilde n}$) after the feature-redistribution in Eq. \ref{eqn:Concatenation} where $\otimes$ is element-wise multiplication and $\oplus$ is element-wise addition.  The final output of ULSAM ($\hat F$) is obtained by concatenating the feature maps from each group (Eq. \ref{eqn:ULSAMFinalOutput}).

Similar to the squeeze operation in SE-Net \cite{Hu_2018_CVPR}, depthwise convolution followed by max pool operation, which highlights the local informative regions \cite{ICLR_2017_PayingMoreAttention}, in Eq. \ref{eqn:AttentionMap} gathers the spatial information. However, unlike the excitation stage in SE-Net, we do not use parameter-heavy MLP to captures the channel-wise inter-dependencies; instead, we exploit the linear relationship between different feature map subspace for integrating the cross-channel information. In effect, ULSAM learns to capture the complex interaction of cross channel information with very few parameters and computations. We analyze ULSAM by considering three prominent cases.\\ 
{\bf Case 1}: $g =1 $: In this case, there is only one group, i.e.,  the cross channel information for the whole feature volume is captured by a single attention map. Intuitively, a single attention map is not sufficient to capture the complex relationships in the entire feature space and would result in lower predictive performance. \\
{\bf Case 2} : $1 < g < m $ : Dividing the feature maps into $g$ groups implies that $g$ attention maps will be generated. Each attention map can capture the cross channel information from the feature maps in its respective groups. We have performed our experiments (Section \ref{sec:Experimental}) with various $g$ (such as $g$= 1,4,8,16) and have obtained performance improvement with higher $g$ ($g=4, 8, 16$) on ImageNet (Table \ref{tab:MV1accuracy}). \\
{\bf Case 3:} $g=m$: Here, the attention map is generated for each feature map in feature space. Hence, Eq. \ref{eqn:AttentionMap} can be re-written as:  
 $   A_{\tilde n}  = softmax(\alpha_2 \otimes (maxpool^{3 \times 3, 1}(\alpha_{1}\otimes F_{\tilde n})))$. Here $\alpha_1$ and $\alpha_2$ are parameters for depthwise and pointwise convolution. In each group, there is only one feature map (i.e., $G = 1$),  and nothing is to be learned along the channel dimension. Hence, the attention map will not be able to capture the cross channel information, and the process of generating attention maps reduces to a non-linear transformation of the feature map itself.

Clearly, better interaction of cross-channel information can be obtained when $1 < g < m$ and this intuition is confirmed by the results shown in Table \ref{tab:MV1accuracy}.  
To keep the model simple, we do not employ any intelligent policy for dividing the feature maps into groups. Instead, we divide the feature maps into groups such that $G$ remains the same in all the groups. We notice that {\em dividing the feature maps into groups does not incur any additional parameter or computation overhead} while learning the cross-channel information effectively.  In other words, the storage and computational cost depends only on the number of channels ($m$) and is independent of the number of groups ($g$) formed.

\section{Experiments and Results} \label{sec:Experimental}

{\bf Datasets:}
We perform image classification on ImageNet-1K \cite{ILSVRC15} and fine-grained datasets, viz.,  Caltech-UCSD Birds-200-2011  \cite{Birds} (Birds for short) and Stanford Dogs \cite{Dogs} (Dogs for short)  datasets (Table \ref{tab:Datasets}). Fine-grained classification is quite challenging due to substantial intra-class variation. Similarly, Food-101  has noisy training samples and high variations in local information. We report validation accuracy (Top-1 and Top-5), an average of three runs, for a single crop input image of size $224\times 224$. The accuracy reported on fine-grained image classification datasets is obtained by training the models from scratch (not on the ImageNet pre-trained model).

\textbf{Experimental setup:} We perform  experiments using PyTorch deep learning framework \cite{paszke2017automatic} with MobileNet-V1 (MV1 for short) and MobileNet-V2 (MV2 for short) as baseline architectures. To enable fair comparison, we reproduce the results of baseline CNNs while applying the same training configuration as in baseline + ULSAM. We used four  P100 GPUs for ImageNet-1K experiments and two P100 GPUs for fine-grained image classification experiments. We train MV1 and MV1+ULSAM with the batch size of 128 and  SGD optimizer with 0.9 momentum for 90 epochs in the case of ImageNet-1K and 150 epochs for fine-grained image classification. The initial learning rate is 0.1, and it reduces to $\frac{1}{10}$th after every 30 epochs.  Similarly, we train MV2 and MV2+ULSAM with batch size 98 and SGD optimizer with 0.9 momentum and 4e-5 weight decay for 400 epochs on ImageNet-1K and 200 epochs for fine-grained image classification. The initial learning rate is set to 0.045 and decays to 0.98$\times$ after every epoch.

\begin{table} [htbp]
\caption{Datasets used in experiments}
\label{tab:Datasets}
\centering 
\resizebox{0.48\textwidth}{!}{
\begin{tabular}{ c| c| c| C{1.5cm}| C{1.5cm}  } 
 \hline
 Dataset & \#Classes & Size (train/test)  & Inter-class variations &  Intra-class variations \\
 \hline
 ImageNet-1K   \cite{ILSVRC15}       &1000 &1.28M/50K   & high & low \\                
 Food-101  \cite{2014_ECCV_Bossard}      &101 &75,750/25,250   & low & high \\ 
  Birds \cite{Birds}          &200 &5,994/5,794  & low & high \\ 
  Dogs \cite{Dogs}     &120 &22K   & low & high \\ 
 \hline
\end{tabular}}
\end{table}

\begin{table*}[htbp]
    \begin{minipage}{0.5\textwidth}
        \centering
        \caption{MV1 architecture}
        \resizebox{0.7\textwidth}{!}{
        \begin{tabular}[htbp]{ c | c | c }
            \hline
           \multicolumn{2}{c}{\bf Layer no. } & { \bf (in, out, stride)} \\
            \hline
            1 & conv2d & ( 3, 32, 2)\\
            \hline
            2,3 & DWS& ( 32,  64, 1), ( 64, 128, 2)\\
            4,5 & DWS & ( 128, 128, 1), ( 128, 256, 2)\\
            6,7 & DWS& ( 256, 256, 1), ( 256, 512, 2)\\
            \hline
            8-12 & DWS & $5\times ( 512, 512, 1)$\\
            \hline
            13, 14 & DWS &( 512, 1024, 2), ( 1024, 1024, 1)\\
            \hline
            \multicolumn{3}{c}{AvgPool, FC, softmax} \\
            \hline
        \end{tabular}}
        
        \label{tab:MV1Architecure}
        
        \vspace{0.5cm}
        \caption{MobileNet-V1 layers 8 to 12 with ULSAM}
        \resizebox{0.65\textwidth}{!}{
        \begin{tabular}[htbp]{ c | c }
            \hline
            {\bf Layer no. } & { \bf Layer type}\\
            \hline
            8& DWS (512, 512, 1)\\
            8:1&ULSAM\\ \hline
            9& DWS (512, 512, 1)\\
            9:1&ULSAM\\ \hline
            10& DWS (512, 512, 1)\\\hline
            11&ULSAM\\ \hline
            12&DWS (512, 512, 1)\\
            \hline
        \end{tabular}}
        
        \label{tab:MobilenetV1withULSAM}
        
    \end{minipage}
    \hfill
    \begin{minipage}{0.5\textwidth}
        \centering
        \caption{MV2 architecture}
        \begin{tabular}[htbp]{ c | c | c }
        
            \hline
           \multicolumn{2}{c}{\bf Layer no./type } & { \bf (in, out) } \\
            \hline
            1 & conv2d & ( 3, 32)\\
            \hline
            2 & residual block & (32, 16)\\
            \hline
            3-4 & residual block & $2\times (16, 24)$ \\
            \hline 
            5 & residual block & ( 24, 32) \\
            6-7 & residual block & $2\times ( 32, 32)$ \\
            \hline 
            8 & residual block & ( 32, 64) \\
            9-11 & residual block & $3\times ( 64, 64)$\\
            \hline 
            12 & residual block & ( 64, 96) \\
            13-14 & residual block & $2\times ( 96, 96)$ \\ 
            \hline 
            15 & residual block & (96, 160) \\
            16-17 & residual block & $ 2\times (160, 160)$ \\
            \hline 
            18  & residual block & (160, 320) \\ 
            \hline 
            19 & conv2d & ( 320, 1280) \\
            \hline
            \multicolumn{3}{c}{AvgPool}   \\
            \hline
            20 & conv2d & (1280, num classes) \\
            
            \hline
        \end{tabular}

        \label{tab:MV2Architecure}
    \end{minipage}
    
\end{table*}

\begin{table*}[htbp] \small
\centering
\caption{Image classification accuracy (\%) of MV1/MV2 (with additional parameters/FLOPs) + ULSAM ($g$ = 1,2,4,8,16) on ImageNet-1K.}
\resizebox{1\textwidth}{!}{
\begin{tabular}{c|c|c|c|c c|c c|c c|c c|c c}
\hline
\multirow{2}{*}{Model} & \multirow{2}{*}{Pos} & \multirow{2}{*}{\#Params} & \multirow{2}{*}{\#FLOPs} &   \multicolumn{2}{c}{$g=1$} & \multicolumn{2}{c}{$g=2$} & \multicolumn{2}{c}{$g=4$} &  \multicolumn{2}{c}{$g=8$} & \multicolumn{2}{c}{$g=16$}\\
 & & & &  Top-1 & Top-5 & Top-1 & Top-5 &  Top-1 & Top-5 & Top-1 & Top-5 & Top-1 & Top-5 \\
\hline
1.0 MV1 (vanilla) & -- & 4.2M & 569M & \multicolumn{8}{c}{Top-1 = 70.65, Top-5 = 89.76}   \\
\hline
1.0 MV1 + ULSAM & 11:1 & 4.2M & 569.2M & 70.69 & 89.85 & 70.84 & 89.87 & 70.77 & {\bf 89.91} & 70.59 & 89.83 & {\bf 70.89} & 89.74 \\
1.0 MV1 + ULSAM & 12:1 & 4.2M & 569.2M & 70.62 & 89.86 & 70.88 & 89.88 & 70.61 & 89.79 & {\bf 70.92} & {\bf 89.98} & 70.73 & 89.78 \\
1.0 MV1 + ULSAM & 13:1 & 4.2M & 569.1M & 70.63 & 89.60 & 70.85 & 89.97 & {\bf 70.86} & 89.85 & 70.74 & 89.81 & 70.82 & {\bf 90.05} \\
\hline
MV2 (vanilla) & -- & 3.4M & 300M & \multicolumn{8}{c}{Top-1 = 71.25, Top-5 = 90.19}   \\
\hline
MV2 + ULSAM & 17:1 & 3.4M & 300.01M & 71.31 & 90.28 & 71.39 & 90.34 & {\bf 71.64} & 90.27 & 71.35 & 90.36 & 71.42 & {\bf 90.43} \\
\hline
\end{tabular}}
\label{tab:MV1accuracy}
\end{table*}

{\bf Where to insert ULSAM in CNNs?} 
Deeper layers have more semantic information and coarser spatial information in feature maps \cite{2019_NIPS_Ramachandran}. Moreover, deeper layers are associated with global features and possess strong positional information \cite{2020_ICLR_Islam}. Hence, applying self-attention in deeper layers better learns the interaction of global information as compared to applying attention in initial layers. Furthermore, since the spatial size of feature maps is smaller (as compared to the feature map size in initial layers),  employing self-attention in deeper layers is computationally cheaper than applying it in the initial layer s\cite{2019_NIPS_Ramachandran}. As shown in Table \ref{tab:MV1Architecure},  MV1  has a stack of 5 layers (layers 8-12) with 512 input and output feature maps. Similarly, MobileNet-V2 (MV2) has residual bottleneck blocks from layers 2 to 18 and has different stacks of either 2 or 3 layers with the same configuration (Table \ref{tab:MV2Architecure}). These blocks, with repeated layers with the same configuration, incurs very high computation overhead due to a higher number of filters. For example, layers 8 to 12 in MV1 cumulatively account for 46\% of total FLOPs in MV1. Therefore, we insert  ULSAM  between the layers and/or it substitutes the layers from the aforementioned blocks to learn the cross-channel interactions efficiently.

\subsection{Results on ImageNet-1K}

{\bf MV1 and MV2 with additional parameters/FLOPs:}
We insert ULSAM in MV1 after layer 11, 12, and 13 which is  represented as 11:1, 12:1, and 13:1 in Table \ref{tab:MV1accuracy}. The parameter and FLOPs overhead incurs due to the insertion of ULSAM at position 11:1, 12:1, and 13:1  is 1.02K and 0.2M/0.1M respectively. Top-1 accuracy on ImageNet-1K has increased by 0.27\% and 0.21\% when position of ULSAM ($g$=4/8) is 12:1 and 13:1, respectively. Similarly, the accuracy of MV2 is increased by 0.39\% when ULSAM is inserted after layer 17 (Table \ref{tab:MV2Architecure}) and incurs only 0.32K and 0.015M additional parameters and  FLOPs respectively (Table \ref{tab:MV1accuracy}).  

{\em Key observations:} For every position of ULSAM, the performance of MV1/MV2 is higher when $g\geq4$. Specifically, there is a significant gain in the performance of MV1/MV2 when $g$ is increased beyond one. This indicates that separate attention maps for the different parts of ofmaps help in better feature representation.

{\bf MV1 and MV2 with fewer parameters/FLOPs:} 
As shown in Table \ref{tab:MV1Architecure}, MV1 layers 8 to 12 have the same configuration and they account for 46\% of total FLOPs. We use ULSAM to optimize this part of the network by inserting ULSAM after $8^{th}$ and $9^{th}$ layers and substitute $11^{th}$ layer, i.e. at position (8:1, 9:1, 11)  shown in Table \ref{tab:MobilenetV1withULSAM}. Compared to baseline network, we achieved a substantial reduction  52M and 0.3M  in FLOPs and parameters (respectively)  with a 0.22\% drop in top-1 accuracy on ImageNet-1K (Table \ref{tab:MV1AblationImageNet}).

\begin{table}[htbp]
\centering
\caption{Image classification accuracy (\%) of MV1 (with fewer parameters and FLOPs) + ULSAM on ImageNet-1K.}
\resizebox{0.48\textwidth}{!}{
\begin{tabular}{c|c|c|c|c|c}
\hline
Models & Pos & \#Params & \#FLOPs & Top-1  & Top-5  \\
\hline
1.0 MV1 (vanilla) & -- & 4.2M & 569M & 70.65 & 89.76 \\
1.0 MV1 + ULSAM ($g$ = 1)& (8:1, 9:1, 11) & 3.9M& 517M  & 69.92 & 89.25 \\
1.0 MV1 + ULSAM ($g$ = 2)& (8:1, 9:1, 11)&  3.9M& 517M &  70.14 & 89.67 \\
1.0 MV1 + ULSAM ($g$ = 4)& (8:1, 9:1, 11)&  3.9M& 517M & {\bf 70.43} & 89.92 \\
1.0 MV1 + ULSAM ($g$ = 8)& (8:1, 9:1, 11) &  3.9M& 517M & 70.29 & 89.96 \\
1.0 MV1 + ULSAM ($g$ = 16)& (8:1, 9:1, 11) & 3.9M& 517M & 70.04 & {\bf 89.98} \\
\hline
0.75 MV1 (vanilla)  & -- & 2.6M& 325M &  67.48 & 88.00 \\
0.75 MV1 + ULSAM ($g$ = 1) & (8:1, 9:1, 11)&  2.4M& 296M & {\bf 67.98} & 88.06 \\
0.75 MV1 + ULSAM ($g$ = 4) & (8:1, 9:1, 11) & 2.4M& 296M  & 67.81 & {\bf 88.43} \\
\hline
0.50 MV1 (vanilla) & -- & 1.3M & 149M &  63.22 & 84.63 \\
0.50 MV1 + ULSAM ($g$ = 1) & (8:1, 9:1, 11) & 1.2M & 136M & {\bf 63.42} & 84.70 \\
0.50 MV1 + ULSAM  ($g$ = 4) & (8:1, 9:1, 11) &  1.2M & 136M & 63.25 & {\bf 84.81} \\
\hline
\end{tabular}}
\label{tab:MV1AblationImageNet}
\end{table}


\begin{table}[htbp] \small
\centering
\caption{Image classification accuracy (\%) of MV2 (with fewer parameters and FLOPs) + ULSAM on ImageNet-1K.}
\resizebox{0.48\textwidth}{!}{
\begin{tabular}{c|c|c|c|c|c}    \hline
 Models & Pos & \#Params &  \#FLOPs & Top-1  & Top-5 \\
\hline
MV2 (Vanilla) & -- & 3.4M  & 300M & 71.25 & 90.19 \\
MV2 + ULSAM ($g$ = 4)& (14, 17) & 2.96M & 261.88M & {\bf 71.52} &  {\bf 90.25}\\
MV2 + ULSAM($g$ = 4)& (16, 17) & 2.77M & 269.07M & 70.74 & 89.15 \\
MV2 + ULSAM ($g$ = 4)& (13, 14, 16, 17) & 2.54M & 223.77M & 69.72 & 87.79 \\
\hline
\end{tabular}}
\label{tab:MV2AblationImageNet}
\end{table}

\begin{table*}[htbp]\small
\centering
\caption{Image classification accuracy (\%) of MV1 (with fewer parameters/FLOPs) + ULSAM ($g$ = 1,4,8,16) on fine-grained datasets.}
\resizebox{0.99\textwidth}{!}{
\begin{tabular}{c|c|c|c|c|c|c|c|c|c}
\hline
\multirow{2}{*}{Models} & \multirow{2}{*}{Pos} & \multirow{2}{*}{\#Params} & \multirow{2}{*}{\#FLOPs} & \multicolumn{2}{c|}{Food-101} & \multicolumn{2}{c|}{Birds} & \multicolumn{2}{c}{Dogs} \\
  & {}  & {} & {} & Top-1 & Top-5 & Top-1 & Top-5 & Top-1 & Top-5 \\ \hline
  
MV1 (vanilla)  & -  & 4.2M &  569M   & 81.31   & 95.24   & 62.88  & 86.05 & 62.20 & 89.66 \\
MV1 +  ULSAM ($g$ = 1) & (8:1, 9:1, 11) & 3.9M & 517M & 81.28 & {\bf 95.50}  & 62.46  & 86.01 & 62.73 & 88.80  \\
MV1 +  ULSAM ($g$ = 4)  & (8:1, 9:1, 11)& 3.9M & 517M & 81.30 & 95.37   & 63.52 & 85.80 & 63.06 & 89.58  \\
MV1 + ULSAM ($g$ = 8) & (8:1, 9:1, 11) & 3.9M & 517M & 81.19  & 95.41  & {\bf 64.44} & \textbf{86.60} & {\bf 63.30} & \textbf{89.68} \\
MV1 + ULSAM ($g$ = 16) & (8:1, 9:1, 11) & 3.9M & 517M & {\bf 81.62} &  95.33 & 63.47 & 84.90& 62.75& 89.35\\    
\hline
\end{tabular}}
\label{tab:MV1FineGrainedAll}
\end{table*}

\begin{table*}[htbp] \small
    \centering
    \caption{Image classification accuracy (\%) of MV2 (with fewer parameters/FLOPs) + ULSAM ($g$ = 1,4,8,16) on Food-101 dataset.}
    \resizebox{0.99\textwidth}{!}{
  \begin{tabular}{c|c|c|c|c c|c c|c c|c c}
    \hline
    \multirow{2}{*}{Model} &
    \multirow{2}{*}{Positions} &
    \multirow{2}{*}{\#Params} &
    \multirow{2}{*}{\#FLOPs} &
      \multicolumn{2}{c}{$g=1$} &
      \multicolumn{2}{c}{$g=4$} &
      \multicolumn{2}{c}{$g=8$} &
      \multicolumn{2}{c}{$g=16$} \\
      & & & &  Top-1 &  Top-5 &  Top-1 &  Top-5 &  Top-1 &  Top-5 &  Top-1 &  Top-5 \\
      \hline
      MV2 (vanilla) & -- & 3.4M & 300M & \multicolumn{8}{c}{Top-1 = 81.51, Top-5 = 95.24}   \\
      \hline
    MV2 + ULSAM & 13 & 3.28M & 277.34M & 81.67 &  {\bf 95.82} &  81.71 & 95.47 &  81.76 & 95.51 & {\bf 81.94} & 95.63 \\
    MV2 + ULSAM & 16 & 3.08M & 284.54M & 82.05 & {\bf 95.56} &{\bf 82.02} & 95.48 &  81.74 & 95.40 & 81.54 & 95.14 \\
    MV2 + ULSAM & (9,13) & 3.23M & 267.06M & 81.66 &  95.36 &  81.72 & 95.48 & {\bf 81.88} & {\bf 95.69} & 81.57 & 95.30 \\
    MV2 + ULSAM & (16,17) & 2.77M & 269.08M & 82.62 & 95.76 & 82.40 & 95.70 & 82.83 & 95.81 & {\bf 83.02} & {\bf 95.87} \\
    MV2 + ULSAM & (14,17) & 2.97M & 261.88M & 81.57 & {\bf 95.44} & 81.69 &  95.36 &  {\bf 82.13} & 95.42 & 81.84 & 95.40 \\
    MV2 + ULSAM & (13,14,16,17) & 2.54M & 224.16M & 82.38 &  95.76 &  82.31 & 95.80 & 82.59 & {\bf 95.82} & {\bf 82.91} & 95.77 \\
\hline
\end{tabular}}
\label{tab:MV2FineGrainedFood}
\end{table*}

\begin{table*}[htbp] \small
    \centering
    \caption{Image classification accuracy (\%) of MV2 (with fewer parameters/FLOPs) + ULSAM ($g$ = 1,4,8,16) on Birds dataset.}
    \resizebox{0.99\textwidth}{!}{
  \begin{tabular}{c|c|c|c|c c|c c|c c|c c}
    \hline
    \multirow{2}{*}{Model} &
    \multirow{2}{*}{Positions} &
    \multirow{2}{*}{\#Params} &
    \multirow{2}{*}{\#FLOPs} &
      \multicolumn{2}{c}{$g=1$} &
      \multicolumn{2}{c}{$g=4$} &
      \multicolumn{2}{c}{$g=8$} &
      \multicolumn{2}{c}{$g=16$} \\
      & & & &  Top-1 &  Top-5 &  Top-1 &  Top-5 &  Top-1 &  Top-5 &  Top-1 &  Top-5 \\
      \hline
      MV2 (vanilla) & -- & 3.4M & 300M & \multicolumn{8}{c}{Top-1 = 62.94, Top-5 = 84.92}   \\
      \hline
      MV2 + ULSAM & 13 & 3.28M & 277.34M & 63.01 & {\bf 85.17} &  63.05 & 83.48 & 63.11 & 84.79 & {\bf 64.32} & 84.62 \\
    MV2 + ULSAM & 16 & 3.08M & 284.54M & 63.98 & {\bf 86.22} & 64.44 & 84.87 &  {\bf 65.03} & 85.63 & 63.47 & 84.45 \\
    MV2 + ULSAM & (9,13) & 3.23M & 267.06M & 63.43 & {\bf 85.55} & {\bf 63.47} & 84.41 & 63.10 & 84.96 & 62.21 & 84.62 \\
    MV2 + ULSAM & (16,17) & 2.77M & 269.08M & 64.19 & 85.46 & 64.57 & 84.92 & 64.61 & 85.25 & {\bf 65.03} & {\bf 85.64} \\
    MV2 + ULSAM & (14,17) & 2.97M & 261.88M & 63.35 & 85.29 & 64.70 & {\bf 86.98} & {\bf 65.41} & 86.01 & 63.31 & 84.92 \\
    MV2 + ULSAM & (13,14,16,17) & 2.54M & 224.16M & 64.11 & {\bf 86.77} & {\bf 64.15} & 84.92 & 63.22 & 85.21 & 63.98 & 85.12 \\
\hline
\end{tabular}}
\label{tab:MV2FineGrainedBirds}
\end{table*}

\begin{table*}[htbp]
    \centering
    \caption{Image classification accuracy (\%) of MV2 (with fewer parameters/FLOPs) + ULSAM ($g$ = 1,4,8,16) on Dogs dataset.}
    \resizebox{0.99\textwidth}{!}{
  \begin{tabular}{c|c|c|c|c c|c c|c c|c c}
    \hline
    \multirow{2}{*}{Model} &
    \multirow{2}{*}{Pos} &
    \multirow{2}{*}{\#Params} &
    \multirow{2}{*}{\#FLOPs} &
    \multicolumn{2}{c}{$g=1$} &
    \multicolumn{2}{c}{$g=4$} &
    \multicolumn{2}{c}{$g=8$} &
    \multicolumn{2}{c}{$g=16$} \\
   & & & &  Top-1 &  Top-5 &  Top-1 &  Top-5 &  Top-1 &  Top-5 &  Top-1 &  Top-5 \\  \hline
MV2 (vanilla) & -- & 3.4M & 300M & \multicolumn{8}{c}{Top-1 = 61.81, Top-5 = 86.88} \\  \hline
MV2 + ULSAM & 13 & 3.28M & 277.34M & 61.63 & 86.58 & 61.80 & 86.68 &  {\bf 62.05} & {\bf 87.30} & 60.31 & 86.63 \\
MV2 + ULSAM & 16 & 3.08M & 284.54M & 62.60 & {\bf 88.20} & 62.90 & 87.30 & {\bf 63.01} & 88.00 & 61.70 & 87.40 \\
MV2 + ULSAM & (9, 13) & 3.23M & 267.06M & 61.73 & 87.08 & 61.73 & 86.86 &  {\bf 62.66} & 87.86 & 62.08 & {\bf 88.11} \\
MV2 + ULSAM & (16, 17) & 2.77M & 269.08M & 63.28 & 88.86 & {\bf 64.30} & {\bf 89.58} & 64.10 & 88.58 & 62.48 & 88.41 \\
MV2 + ULSAM & (14, 17) & 2.97M & 261.88M & 62.86 & 87.88 & 62.50 & 88.03 & {\bf 64.33} & {\bf 89.31} & 61.67 & 87.18 \\
MV2 + ULSAM & (13, 14, 16, 17) & 2.54M & 224.16M & 63.20 & {\bf 88.84} & {\bf 63.53} & 88.63 & 62.75 & 88.18 & 62.50 & 88.56 \\

\hline
\end{tabular}}  
\label{tab:MV2FineGrainedDogs}
\end{table*}

We further perform experiments on scaled version of MV1 where the number of filters in each layer is scaled down by a factor of $\alpha$ (where $\alpha$ $\in$ \{0.5, 0.75\}).  The position of ULSAM is same  i.e., (8:1, 9:1, 11) as employed in 1.0-MV1+ULSAM. Since performance of MV1 with ULSAM is highest for $g$=4 (Table \ref{tab:MV1AblationImageNet}), we  perform our experiments with only $g$ = 1 and $g$ = 4 for scaled MV1. Table \ref{tab:MV1AblationImageNet} shows the results. {\em Interestingly}, scaled MV1+ULSAM (with both $g$=1 and $g$=4) achieved higher Top-1 accuracy  on ImageNet-1K with significantly reduced parameters and FLOPs compared to their baseline. More precisely, the top-1 accuracy of  0.75-MV1+ULSAM($g$=1) is improved by 0.5\% with 9.1\% and 5.8\% fewer FLOPs and parameters while that of  0.50-MV1+ULSAM($g$=1) is improved by 0.10\% with 8.92\% and 7.69\% fewer  FLOPs and parameters, respectively.

Similarly,  when ULSAM substituted with residual block 14 and 17 in MV2 (Table \ref{tab:MV2Architecure}) the top-1 accuracy of MV2 is improved by 0.27\% while having 431.6K and 38.2M fewer parameters and FLOPs (Table \ref{tab:MV2AblationImageNet}). On substituting layers 13, 14, 16, and 17 with ULSAM, we achieved 25.28\% and 25.27\% reduction in FLOPs and parameter count, respectively, and the top-1 accuracy is reduced to 69.72\%. Thus, {\em ULSAM either retains or improves accuracy while bringing a significant reduction in the number of parameters and FLOPs}.

{\em Key observations:}
\begin{itemize}
\item Unlike MV1, the performance of scaled-MV1 and MV2 is increased with fewer parameters and FLOPs as compared to their baseline networks. This implies that ULSAM exploits spatial and channel redundancy in a batter when the network is more compact. That is, ULSAM captures better inter-class variation through the interaction of cross-channel information when channel redundancy is low.
\item With increasing $g$, there is a diminishing return in the accuracy of MV1+ULSAM (Table \ref{tab:MV1AblationImageNet}) because increasing $g$ reduces the value of $G$ in each group and inhibits the cross-channel information exchange. Also, the top-1 accuracy (70.43\%) of MV1+ULSAM on ImageNet-1k is lower. However,  the top-5 accuracy is higher than that of the baseline (except with $g=1$). This implies that the misclassification,  which leads to lower top-1 accuracy, is correctly predicted in top-5 predictions by the MV1+ULSAM.
\item For both  0.75-MV1+ULSAM and 0.50-MV1+ULSAM, the highest top-1 accuracy is achieved with $g=1$ instead of the $g=4$. Since the scaled versions of MV1 already have fewer feature maps per layer, further dividing the feature maps into groups reduces $G$ (as happened with  $g = 4$) and hence, reduces the information present in each group and degrades accuracy. The top-5 accuracy is highest at $g=4$.
\end{itemize}

\subsection{Results on Fine-grained Image Classification Datasets}

We perform our experiments on three fine-grained datasets (Table\ref{tab:Datasets}) with $g$=1, 4, 8, and 16. \\
{\bf MobileNet-V1 + ULSAM:}
The Results for MV1+ULSAM on fine-grained datasets are shown in Table \ref{tab:MV1FineGrainedAll}. The top-1 accuracy of MV1+ULSAM is improved by 0.31\% on Food-101 with $g$=16, 1.56\% on Birds dataset with $g$=8, and 1.10\% with $g$=8 with 9.14\% and 7.14\% reduction in FLOPs and parameters compared to baseline networks. The highest top-1 accuracy on Birds and Dogs dataset is achieved with $g$ = 8 whereas that on Food-101 is achieved with $g$ = 16 (Table \ref{tab:MV1AblationImageNet}). Thus, at higher value of  $g$, ULSAM  captures the intra-class variation more effectively.

{\bf MobileNet-V2 + ULSAM:} 
The  experimental results on Food-101, Birds and Dogs datasets  are shown in Table \ref{tab:MV2FineGrainedFood}, Table \ref{tab:MV2FineGrainedBirds}, and Table \ref{tab:MV2FineGrainedDogs} respectively. The performance of MV2 has significantly improved with a substantial reduction in parameter and FLOPs counts.  For example,  at position (13,14,16,17) the top-1 accuracy of MV2 is improved by 1.4\% on Food-101 (with $g$=16), 1.21\% on Birds dataset (with $g$=4), and 1.72\% on Dogs dataset (with $g$=4) while incurring 25.28\% and 25.27\% reduction in FLOPs and parameter count, respectively. On  all positions of ULSAM, the performance is improved compared to baseline model.

{\em Key Observations:} 
Almost all experiments on fine-grain datasets with different positions of ULSAM perform better than vanilla with significantly lower computations and parameters. This substantiates that applying different weights (through different attention maps) for feature maps with different frequency components boost the representational power significantly on fine-grained datasets.

\subsection{Attention visualization}
\begin{figure}[htbp]
    \centering
    \resizebox{0.5\textwidth}{!}{
    \begin{tabular}{ccccc}
      \small{Input image} &  \small{MV1} &  \small{MV1+ULSAM} &   \small{MV2} &  \small{MV2+ULSAM} \\

    \includegraphics[height=0.85in]{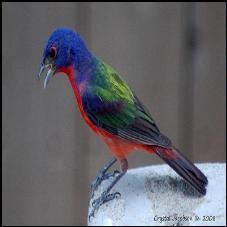} &
    \includegraphics[height=0.85in]{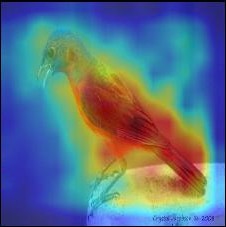} & 
    \includegraphics[height=0.85in]{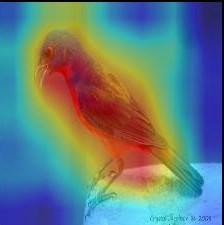} &
    \includegraphics[height=0.85in]{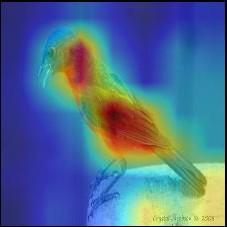} &
    \includegraphics[height=0.85in]{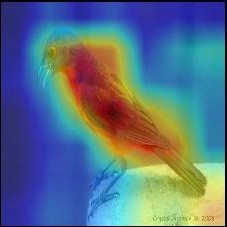} \\
    
    \includegraphics[height=0.85in]{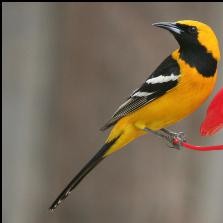} &
     \includegraphics[height=0.85in]{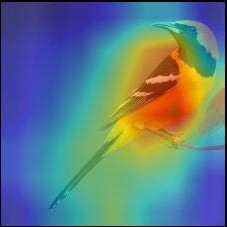} & 
    \includegraphics[height=0.85in]{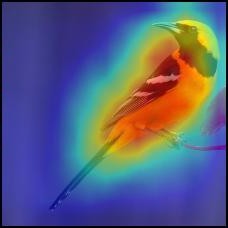} &
    \includegraphics[height=0.85in]{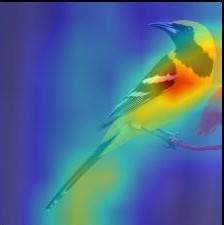} &
    \includegraphics[height=0.85in]{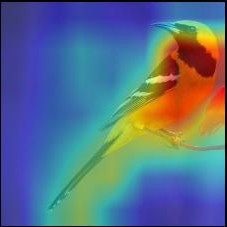} \\
    
     \includegraphics[height=0.85in]{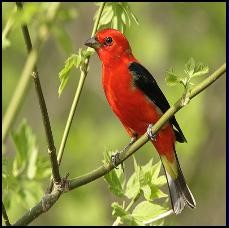} &
     \includegraphics[height=0.85in]{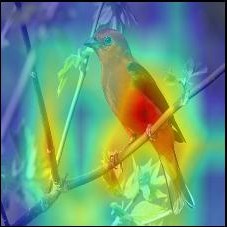} & 
    \includegraphics[height=0.85in]{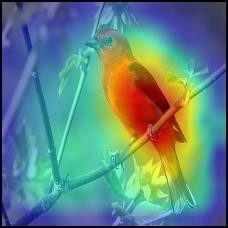} &
    \includegraphics[height=0.85in]{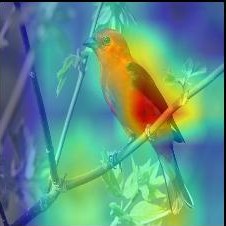} &
    \includegraphics[height=0.85in]{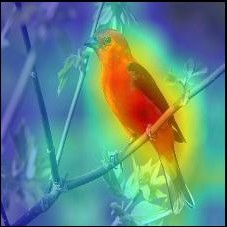} \\
    
     \includegraphics[height=0.85in]{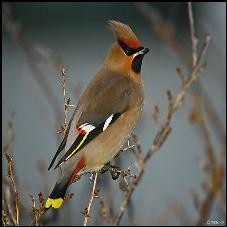} &
     \includegraphics[height=0.85in]{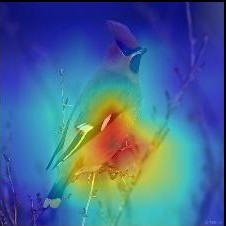} & 
    \includegraphics[height=0.85in]{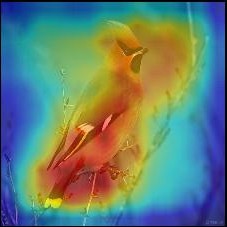} &
    \includegraphics[height=0.85in]{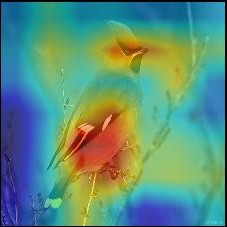} &
    \includegraphics[height=0.85in]{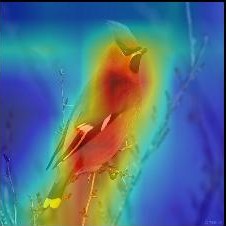} \\

\end{tabular}
}
\caption{{\bf Grad-CAM++ visualization results.} Here the input images are Painted bunting,  Hooded Oriole, Scarlet Tanager, and Bohemian Waxwing (in order). }
\label{fig:Visualization}    
\end{figure}

We now show the effectiveness of our approach through human-interpretable visual explanation. We apply the Grad-Cam++  tool \cite{2018_WACV_GradCam++,ICCV_2017_GradCam} on MV1 and MV2 using images from the Birds dataset. Grad-Cam++ provides complete heatmaps of the object, and the masked regions in the images are important (considered by the networks) for predicting the class. 
Figure \ref{fig:Visualization} shows the visualization results with vanilla MV1 and MV2 and their ULSAM integrated versions. Evidently, ULSAM integrated versions of MobileNet focuses the target images better than the baseline networks, which better explain the effectiveness of ULSAM.

\section{Conclusion} 
We proposed an ultra-lightweight attention block (ULSAM), which divides the feature maps into multiple subspaces to capture the inter-dependencies. The lower parameter and computation overhead of ULSAM, compared to the state-of-the-art attention mechanism, make it desirable for compact CNNs.
Our future work will focus on incorporating spatial attention and capturing complex global relations in the entire feature space of CNNs. 

\section*{Acknowledgments}

We thank anonymous reviewers for their invaluable feedback and insightful reviews.  We thank TRI-AD for the travel grant. This project was supported in part by Semiconductor research corporation (SRC) and by Science and Engineering Research Board (SERB), India, award number ECR/2017/000622.

{\small
\bibliographystyle{ieee}
\bibliography{Reference}
}

\end{document}